\documentclass[review]{elsarticle}

\usepackage[english]{babel}
\usepackage{graphicx}
\usepackage{url}
\usepackage{hyperref}

\usepackage{amssymb}
\usepackage{amsthm}
\usepackage{amsfonts}
\usepackage{amsmath}
\usepackage{xcolor}

\usepackage{graphicx}
\usepackage{psfrag}
\usepackage{adjustbox}
\usepackage{multirow}
\usepackage{lineno}
\usepackage{booktabs} 
\usepackage{siunitx} 

\usepackage{float}
\usepackage[linesnumbered,ruled,vlined]{algorithm2e}

\usepackage[normalem]{ulem}

\journal{Computers and Electronics in Agriculture}

\newcommand{\T}{{\mathrm{T}}}
\newcommand{\Tr}{{\mathrm{Tr}}}
\newcommand{\R}{{\mathrm{R}}}

\begin{document}

\begin{frontmatter}

\title{Advanced wood species identification based on multiple anatomical sections
and using deep feature transfer and fusion}

\author[1]{Kallil M. Zielinski}
\author[1]{Leonardo Scabini}
\author[2]{Lucas C. Ribas}
\author[3]{Núbia R. da Silva}
\author[4]{Hans Beeckman}
\author[5]{Jan Verwaeren}
\author[1,6]{Odemir M. Bruno}
\author[6]{Bernard De Baets}

\address[1]{\small{S\~{a}o Carlos Institute of Physics, University of S\~{a}o Paulo, Brazil}}
\address[2]{\small{Institute of Biosciences, Humanities and Exact Sciences, São Paulo State University, Brazil}}
\address[3]{\small{Federal University of Catalão, Brazil}}
\address[4]{\small{Royal Museum for Central Africa, Belgium}}
\address[5]{\small{BIOVISM, Dept.\ of Data Analysis and Mathematical Modelling, Ghent University, Belgium}}
\address[6]{\small{KERMIT, Dept.\ of Data Analysis and Mathematical Modelling, Ghent University, Belgium}}

\begin{abstract}
In recent years, we have seen many advancements in wood species identification. Methods like DNA analysis, Near Infrared (NIR) spectroscopy, and Direct Analysis in Real Time (DART) mass spectrometry complement the long-established wood anatomical assessment of cell and tissue morphology. However, most of these methods have some limitations such as high costs, the need for skilled experts for data interpretation, and the lack of good datasets for professional reference. Therefore, most of these methods, and certainly the wood anatomical assessment, may benefit from tools based on
Artificial Intelligence. In this paper, we apply two transfer learning techniques with Convolutional Neural Networks (CNNs) to a multi-view Congolese wood species dataset including sections from different orientations and viewed at different microscopic magnifications. We explore two feature extraction methods in detail, namely Global Average Pooling (GAP) and Random Encoding of Aggregated Deep Activation Maps 
(RADAM), for efficient and accurate wood species identification. Our results indicate superior accuracy on diverse datasets and anatomical sections, surpassing the results of other methods. Our proposal represents a significant advancement in wood species identification, offering a robust tool to support the conservation of forest ecosystems and promote sustainable forestry practices.

\end{abstract}


\end{frontmatter}


\section{Introduction}\label{sec:intro}
Wood is a versatile and renewable resource that can be produced in a sustainable way.
It is widely used in many industries such as construction, furniture and paper production. The global demand for wood has led to the emergence of illegal logging and trade, having environmental, social, and economic repercussions. Illegal wood trade represents a significant portion of global wood trade, with percentages increasing in regions such as Southeast Asia, Central Africa, and South America~\cite{May2017TransnationalWorld}. This illicit trade, worth billions of dollars annually, also threatens ecosystems due to the over-exploitation of rare and protected species. To combat this issue, various protection measures, such as the Convention on International Trade in Endangered Species of Wild Fauna and Flora (CITES)~\cite{CITESConventionFlora}, and policy measures like the European Union Timber Regulation (EUTR) and the U.S. Lacey Act have been 
implemented~\cite{ITTO2020BiennialSituation}.

The effective implementation of the above policies and regulations also requires efficient methods for identifying wood species as well as robust datasets. Currently, wood species identification is primarily done through wood anatomical analysis, which involves the examination of tissue and cell diagnostic features using various imaging tools such as hand lenses, light or electronic microscopes, 2D and 3D scans, among others. Also, the International Association of Wood Anatomists (IAWA) has developed a list of standardized microscopic diagnostic features that can be used to identify hardwood species based on anatomical patterns, such as vessels, rays, parenchyma, and fibers~\cite{Wheeler1989IAWAIdentification}. Although this approach is widely applied, readily available, and cost-effective, it can sometimes fail to distinguish between closely related taxa or determine the exact species~\cite{Dormontt2015ForensicLogging,Gasson2011HowCites}. 

Alternative methods for wood species identification have been gradually developed, including DNA analysis, Near Infrared spectroscopy, and Direct Analysis in Real Time (DART) mass spectrometry~\cite{Braga2011TheII,Hassold2016DNASpecies,Pastore2011NearCurupixa,Price2021PterocarpusSpectrometry,Jiao2018DNASpecimens}. These methods show promising results but are still hindered by factors such as high costs, the need for skilled experts for data interpretation, and the lack of reference datasets. Recently, pattern recognition techniques employing machine vision for automated wood species identification have emerged as a feasible and attractive solution. This approach is less dependent on expert knowledge and can leverage existing datasets containing high-quality microscopy images~\cite{Nithaniyal2014DNAIndia, Hanssen2011MolecularPrimers}. 

The state-of-the-art in wood species identification has seen significant progress through the incorporation of automated classification techniques based on macroscopic and microscopic images. Several 
studies~\cite{Bremananth2009WoodSystem, Zhao2014WoodScheme, Guang-Sheng2013WoodAnalysis, Khalid2011TropicalClassifiers} have indeed reported promising results. 
However, these studies either focus on a limited number of species or rely on morphological wood features, which are dependent on segmentation and may consequently yield variable outcomes.

Texture analysis has emerged as a promising technique, as it can describe the spatial organization of pixels and the variation of patterns in an area on the surface of the studied object. Filho 
et al.~\cite{Filho2014ForestImages} and Wang et al.~\cite{Wang2013ARecognition} used texture attributes derived from macroscopic images for wood species identification. Martins et al.~\cite{Martins2012CombiningRecognition} and Cavalin et al.~\cite{Cavalin2013ARecognition} employed texture features to identify wood species from the Brazilian flora using microscopic transverse cross-sections. These studies made use of Local Phase Quantization (LPQ), Local Binary Patterns (LBP) and gray-level co-occurence matrices as feature descriptors.

Various computer vision models have been employed to automate wood species identification using digital imagery of anatomical sections. These models typically involve collecting a representative dataset of labeled digital images, applying a feature extraction procedure, and training a machine learning classification algorithm. Martins et al.~\cite{Martins2013ASpecies} achieved an accuracy of 86\% using LBP as a feature descriptor combined with Support Vector Machines (SVMs). Filho 
et al.~\cite{Filho2014ForestImages} used a strategy of dividing the image into sub-images, classifying them independently using SVMs, and fusing the class probabilities through a fusion rule, achieving an accuracy of 97.77\% for 41 different species.

Recent studies have also employed deep convolutional neural networks (CNNs) for wood species identification. Ravindran et al.~\cite{Ravindran2018ClassificationNetworks} obtained an accuracy of 87.4\% using CNNs on a dataset of 2303 macroscopic images from the Meliaceae family. Another work~\cite{Ravindran2020TheProducts} proposed a different CNN model to identify 12 common species in the United States based on macroscopic imagery of transverse sections reaching an accuracy of 97.7\%. Lens et al.~\cite{Lens2020Computer-assistedSections} proposed yet another CNN model achieving a similar accuracy of over 98\% on 2240 images from 112 species. 

Transfer learning techniques have recently emerged as a promising approach to image classification in various domains, including wood species identification. Transfer learning enables the application of knowledge gained from one task to a different but related task. It has been particularly successful in situations where the target task has only limited labeled data available, by leveraging pre-trained backbones on larger, more diverse datasets~\cite{Pan2010ALearning}. In the context of wood species identification, transfer learning can be used to fine-tune pre-trained backbones, such as CNNs, to effectively learn the intricate features and patterns of wood species from the available datasets~\cite{Tajbakhsh2020EmbracingSegmentation}.

The application of transfer learning to wood species identification can further improve the efficiency and accuracy of classification models, especially when combined with the aforementioned methods such as LBP, SVMs, and CNNs. For example, Zhao et al.~\cite{Zhao2019DeepMonitoring} employed transfer learning with a pre-trained CNN model for wood species identification, achieving an accuracy of 95.2\% on a dataset of 1832 macroscopic images from 32 species. This approach not only yielded impressive classification results but also reduced the time and computational resources required for training the model.

Despite the significant advancements in wood species identification, the African continent, particularly the Congo Basin, remains underrepresented. To address this gap, we leverage the commercial timber species dataset from the Democratic Republic of Congo (DRC) provided by the Belgian Royal Museum for Central Africa (RMCA)~\cite{congoDB}, focusing on texture features extracted from distinct microscopic cross-sections, which have shown notable discriminative potential in wood classification in works by Rosa da Silva et al.~\cite{RosadaSilva2017AutomatedSpecies, daSilva2022ImprovedPlanes}. To further advance the field of wood species identification, our work applies transfer learning to the timber species dataset from the DRC. This dataset includes 805 images representing 77 distinct wood species across three anatomical sections: transversal, tangential, and radial. In their approach, 
Rosa da Silva et al.~\cite{daSilva2022ImprovedPlanes} employed the LPQ method~\cite{lpq} on each anatomical section, followed by a separate classification using a random forest~\cite{randomforest} algorithm, and a subsequent concatenation of the probability matrices to feed a logistic regression model. Building upon this foundation, we propose two novel transfer learning approaches that harness the power of pre-trained backbones and advanced learning techniques, aiming to enhance the identification efficiency and accuracy of wood species.

The first approach employs a serial activation map fusion using Global Average Pooling (GAP) on a backbone CNN model that has been pre-trained on ImageNet, a large-scale image classification dataset containing millions of samples spanning thousands of object categories~\cite{Deng2009ImageNet:Database}. This allows us to benefit from a richer set of feature representations learned on diverse images.

The second approach involves the use of RADAM (Random encoding of Aggregated Deep Activation Maps), a feature extractor based on randomized auto-encoders that has demonstrated state-of-the-art accuracy on texture recognition tasks~\cite{scabini2023}. By incorporating RADAM into our wood species identification pipeline, we aim to further optimize the classification performance by exploiting the advanced learning capabilities of this approach.

Through the implementation of the above approaches, we seek to improve upon the results achieved by 
Rosa da Silva et al.~\cite{daSilva2022ImprovedPlanes} and contribute to the development of more efficient and more accurate wood species identification models, ultimately promoting the protection and conservation of forest ecosystems.

\section{Materials and Methods} \label{sec:materials}

This section presents the materials and methods used in our study, including the dataset, the feature extraction methods adopted and the parameters of the experimental configurations employed. The following subsections provide a detailed overview of these components.

\subsection{Datasets} \label{sec:datasets}

\begin{figure}[!htb]
    \psfrag{TVS}{\hspace{-4.2em}\scriptsize{Transversal Semi-thin Section}}
    \psfrag{TGS}{\hspace{-4.2em}\scriptsize{Tangential Semi-thin Section}}
    \psfrag{RS}{\hspace{-4.2em}\scriptsize{Radial Semi-thin Section}}
    \centering
    \includegraphics[width=\textwidth]{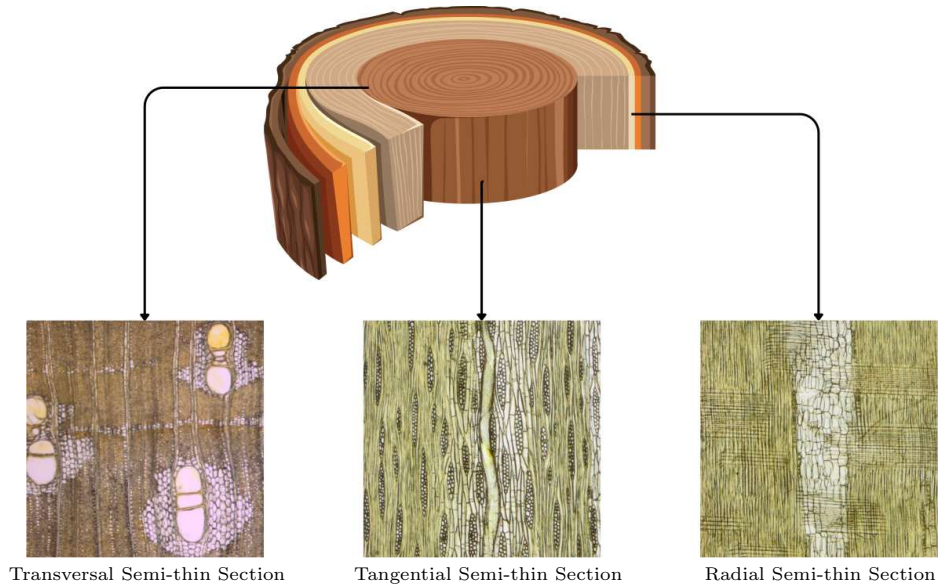}
    \caption{Representation of each anatomical section of a wood sample from the species \textit{Afzelia africana}. The arrows point from the respective regions on the sample (transversal, tangential, and radial) to their corresponding images. Each image offers unique structural characteristics that are crucial for wood species identification. 
    }
    \label{fig:tree_sections}
\end{figure}

Our research uses the timber species dataset from the DRC, which was assembled by Rosa da Silva
et al.~\cite{RosadaSilva2017AutomatedSpecies}. This dataset encompasses a total of 78 different Congolese timber species, each represented by a set of high-resolution images characterized by a resolution of $1000\times1000$ pixels. The uniqueness of this dataset lies in its anatomical diversity. For every timber species included, the dataset provides images from three distinct anatomical planes: tangential, transversal and radial. This approach offers a comprehensive view of each species, facilitating a more accurate identification process. In summary, the dataset is composed of 805 images for each anatomical plane, resulting in a significant volume of data for our method.
Figure~\ref{fig:tree_sections} displays an illustrative representation of the images from each anatomical section in this dataset, which shows the different visual characteristics present in each anatomical plane.

The dataset contains an average of 10 images per class, which is a relatively low number for machine learning tasks. To overcome this limitation, Rosa da Silva et al.~\cite{daSilva2022ImprovedPlanes} 
applied some data augmentation techniques to increase the number of images per class. Starting from the original dataset consisting of $1000\times1000$ pixel images, they created three additional datasets, each with varying image dimensions and based on different transformations. The procedures leading to the different sub-datasets are detailed below:
\begin{itemize}
    \item \textbf{$1000 \times 1000$ (Original)}: This dataset is referred to as the `original', because it contains the initial images as previously described. It includes 805 images for each of the three anatomical sections, with each image consisting of $1000\times1000$ pixels;
    \item \textbf{$1000\times500$}: This dataset is derived from the original by dividing each image into two equal halves, effectively doubling the total number of images compared to the original dataset;
    \item \textbf{$500\times500$}: Each image in the $1000\times 500$ dataset is divided once more, resulting in four separate image sections from each original image in the $1000\times 1000$ dataset;
    \item \textbf{$500\times500$-OGRN}: This dataset is equivalent to the $500\times500$ dataset in terms of image subdivision but also includes the results of several transformations applied to each quarter section of the original images. The first quarter remains identical to the original (O), the second quarter undergoes a Gaussian smoothing operation (G), the third one is rotated by 90 degrees (R), and the final one comes with added salt-and-pepper noise with a density of 0.05~(N). 
\end{itemize}

\begin{figure}[!htpb]
    \psfrag{A}[][]{\scriptsize{a) 1000 $\times$ 1000 (original)}}
    \psfrag{B}[][]{\scriptsize{b) 1000 $\times$ 500}}
    \psfrag{C}[][]{\scriptsize{c) 500 $\times$ 500}}
    \psfrag{D}[][]{\scriptsize{d) 500 $\times$ 500 - OGRN}}
    \centering
    \includegraphics[width=\textwidth]{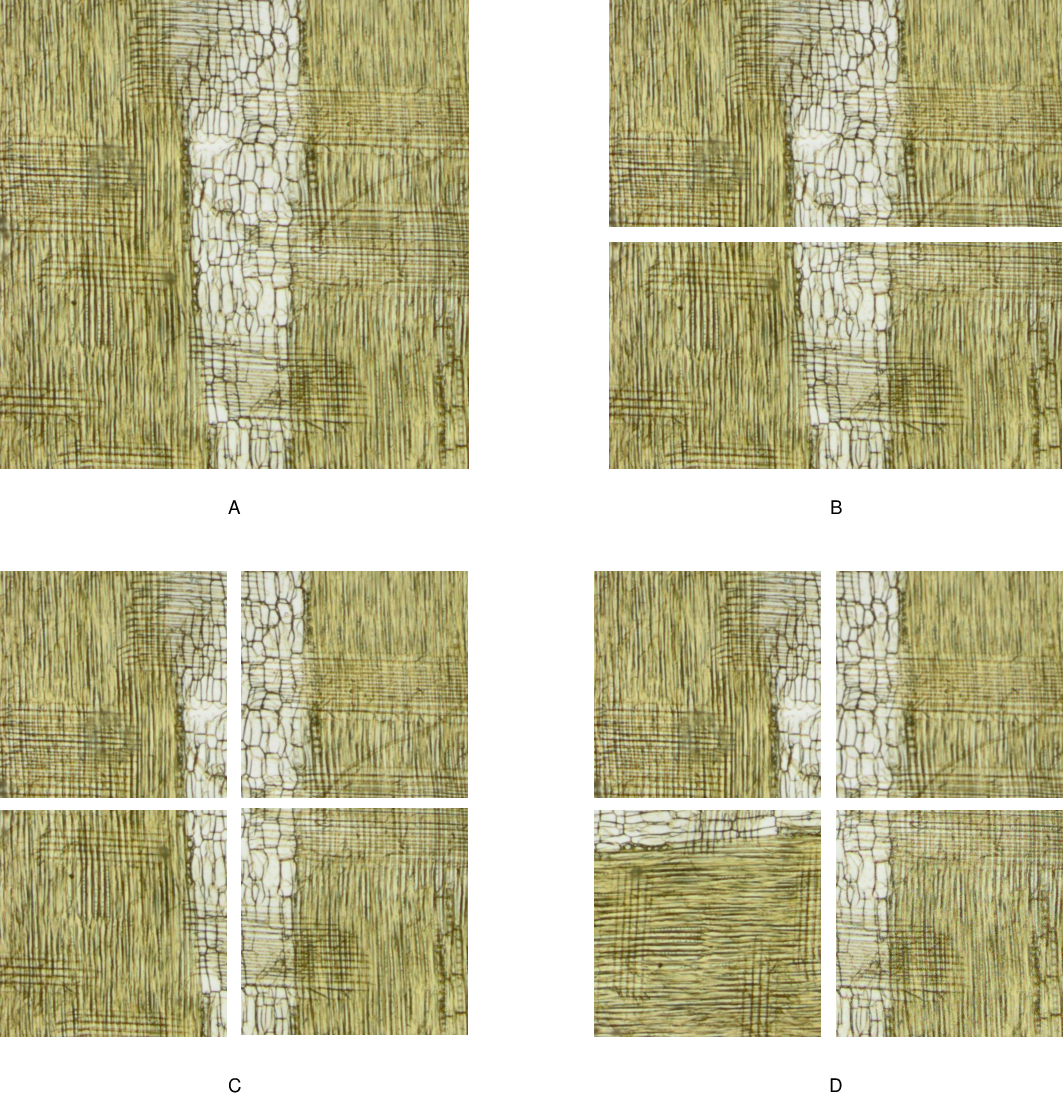}
    \caption{Illustration of the data augmentation techniques employed for each sub-dataset. (a) 1000 $\times$ 1000 original dataset, (b) 1000 $\times$ 500 dataset with each image halved, (c) 500 $\times$ 500 dataset with each original image split into four equal parts, and (d) 500 $\times$ 500 - OGRN dataset demonstrating the applied transformations including original (O), Gaussian smoothing (G), 90-degree rotation (R), and salt-and-pepper noise addition (N).}
    \label{fig:dataaug}
\end{figure}

These additional datasets created from the original set of 1000 $\times$ 1000 images not only increase the quantity of training data but also introduce a range of variations that improve the robustness of the machine learning models trained on them. These data augmentation processes are visually demonstrated in Figure~\ref{fig:dataaug}, where each subfigure corresponds to a different dataset created from the original set of 1000 $\times$ 1000 images. This visualization allows for a better understanding of the transformations applied.

\subsection{Feature Extraction Methods} \label{sec:featureExtraction}

Texture analysis using CNNs typically adopts end-to-end models that leverage a pre-trained backbone, followed by fine-tuning of the architecture specific to the texture recognition problem at hand. In contrast, our study employs two approaches that use the backbone only as a feature extractor with a dedicated classifier applied to the resulting features. Notably, this process eliminates the need for fine-tuning of the architecture.


\subsubsection{GAP}

The first approach we consider is using GAP computed over the activation maps produced by the last layer of the backbone, which are subsequently fed to an SVM classifier~\cite{Cortes1995Support-vectorNetworks}.
Let us denote the function representing the pre-trained backbone as $f$, and the input image from each of the anatomical planes (tangential, transversal, and radial) as $X_\T$, $X_\Tr$, and $X_\R$,
respectively, with $X_\T, X_\Tr, X_\R \in \mathbb{R}^{w_0\times h_0 \times 3}$. We extract features from the last layer of the backbone as follows:
\begin{equation}
    Z_{\T} = f(X_\T), \quad Z_{\Tr} = f(X_{\Tr}), \quad Z_\R = f(X_\R)\,.
\end{equation}

For each wood sample, we have three feature vectors corresponding to the three anatomical planes. We can combine these features in two ways:
\begin{enumerate}
    \item \textbf{Serial Feature Concatenation (SFC)}: We concatenate the feature vectors end-to-end. This can be defined mathematically as:
    \begin{equation}
     Z_{s} = [Z_\T, Z_{\Tr}, Z_\R]\,.
     \label{eq:sfc}
    \end{equation}
    \item \textbf{Parallel Feature Merging (PFM)}: We perform an element-wise summation of the feature vectors:
    \begin{equation}
        Z_{p} = Z_\T \oplus Z_{\Tr} \oplus Z_\R\,.
        \label{eq:pfm}
    \end{equation}
\end{enumerate}

Finally, we use the combined features ($Z_{s}$ or $Z_{p}$) as input to an SVM classifier. Let $h$ denote the SVM, then 
the predicted class label is given by:
\begin{equation}
    \hat{y} = h(Z_{s}) \quad \text{or} \quad \hat{y} = h(Z_{p})\,.
    \label{eq:classification}
\end{equation}

\subsubsection{RADAM}

The second approach we adopt is RADAM~\cite{scabini2023}, which significantly differs
from GAP. Instead of considering only the last layer of a pre-trained backbone, RADAM takes into account multiple activation maps at different depths in the architecture. This approach allows to capture various levels of texture characteristics, from simple to complex features. 

In the RADAM method, the features are combined through a specific sequence of steps, including a Randomized Autoencoder (RAE)~\cite{kasun2013representational,cambria2013extreme}. A RAE is a type of neural network that consists of one hidden layer and aims to reproduce its input as target output, i.e., $Y = X$. The input weights of the RAE are randomly generated using a Linear Congruential Generator (LCG)~\cite{Knuth1997TheAlgorithms}. After training the RAE using the least-squares method, the output weights are used as features in our approach, as we describe below.  

Let us now describe each step of the RADAM method. We denote the function representing the pre-trained backbone as $f$, and its intermediate activation maps at different depths $i$ as $f_i$. Each depth level corresponds to the output of a specific convolutional block in the backbone. For each input image of the three anatomical planes ($X_\T$, $X_{\Tr}$ and $X_\R$) and each depth~$i$, we extract the corresponding activation map:
\begin{equation}
    A_{\T,i} = f_i(X_\T), \quad A_{\Tr,i} = f_i(X_{\Tr}), \quad A_{\R,i} = f_i(X_\R)\,.
\end{equation}

Next, we compute the depthwise $2p$-norm of the activation maps. Let us denote the depthwise norm function as $g$, which takes the activation map as input and applies the $2p$-norm along the depth (channel) dimension. The computed norms for the three anatomical planes at depth $i$ are given by:
\begin{equation}
N_{\T,i} = g(A_{\T,i}), \quad N_{\Tr,i} = g(A_{\Tr,i}), \quad N_{\R,i} = g(A_{\R,i})\,.
\end{equation}
The computed norms are then concatenated along the third dimension (channel dimension $z_i$). However, since the spatial dimensions ($w_i$ and $h_i$) of the activation maps might differ across different depths, we resize the spatial dimension of all activation maps to the same spatial size using bilinear interpolation. The concatenations of the normalized activation maps for the three anatomical planes are denoted as:
\begin{equation}
C_\T = \bigoplus_{i=1}^{n} N_{\T,i}, \quad C_{\Tr} = \bigoplus_{i=1}^{n} N_{\Tr,i}, \quad C_\R = \bigoplus_{i=1}^{n} N_{\R,i}\,,
\end{equation}
where $\bigoplus$ denotes the concatenation operation along the third dimension for each computed anatomical section norm, with $i$ varying from $1$ to $n$, and $n$ representing the number of depths considered in the RADAM approach. 

Next, we apply a set of $m$ RAEs to the concatenated activation maps $C_{\T}$, $C_{\Tr}$ and $C_{\R}$. The projection obtained by the output weights of each RAE is considered as the encoded representation of the activation map. This representation is denoted as $r_j$ for the $j$-th RAE:
\begin{equation}
Z_{\T,i} = r_j(C_{\T}), \quad Z_{\Tr,j} = r_j(C_{\Tr}), \quad Z_{\R,j} = r_j(C_{\R})\,.
\end{equation}
After repeating this process for $m$ RAEs, the representations are combined using the PFM strategy, summating the features for each anatomical section image:
\begin{equation}
    Z_{\T} = \sum_{j=1}^{m} Z_{\T,j}, \quad Z_{\Tr} = \sum_{j=1}^{m} Z_{\Tr,j}, \quad Z_{\R} = \sum_{j=1}^{m} Z_{\R,j}\,.
\end{equation}

In a last step before inputting these features into a classifier, we combine the features of all three anatomical sections using the SFC or PFM strategy just as for GAP (see Eqs.~\eqref{eq:sfc}) and \eqref{eq:pfm}).
Finally, these combined features serve as input to an SVM classifier, which performs the final classification task.

\begin{figure}[!htpb]
    \psfrag{TVSI}[][]{\scriptsize{Transversal Section Image}}
    \psfrag{TGSI}[][]{\scriptsize{Tangential Section Image}}
    \psfrag{RSI}[][]{\scriptsize{Radial Section Image}}
    \psfrag{PTB}[][]{\tiny\parbox{2cm}{\centering Pre-trained \\ Backbone}}
    \psfrag{RF}[][]{\tiny\parbox{2cm}{\centering Radial \\ Features}}
    \psfrag{TGF}[][]{\tiny\parbox{2cm}{\centering Tangential \\ Features}}
    \psfrag{TVF}[][]{\tiny\parbox{2cm}{\centering Transversal \\ Features}}
    \psfrag{AGG}[][cb]{\tiny\parbox{2cm}{\centering Feature \\ Fusion}}
    \psfrag{CL}[][ct]{\scriptsize\parbox{5cm}{\centering SVM Classifier}}
    \psfrag{OUT}[][]{\tiny\parbox{2cm}{\centering Model \\ Output}}
    \centering
    \includegraphics[width=\textwidth]{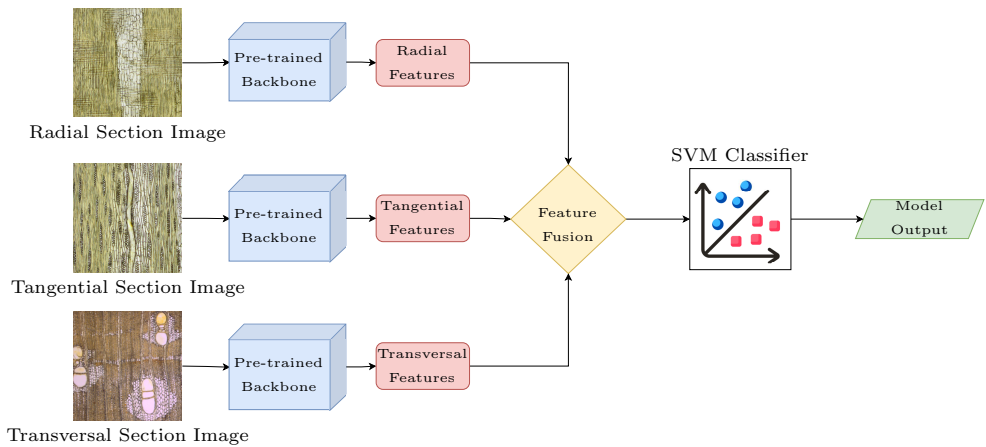}
    \caption{Methodological process employed in this study. Anatomical sections are input to a pre-trained backbone to extract feature blocks. These blocks are then fused and classified using an SVM to produce the final wood species prediction.}
    \label{fig:methodology}
\end{figure}

In order to facilitate the understanding of the overall process, Figure~\ref{fig:methodology} shows the methodology adopted in this paper. The individual anatomical sections (radial, tangential and transversal) are first input through the pre-trained backbone, resulting in their respective feature blocks. These features are then combined through SFC or PFM, forming a unified feature representation. This feature representation is then passed through an SVM classifier for final wood species classification. The output of the model is the predicted wood species, derived from the combined and classified features of the three anatomical sections.

\subsection{Experimental Configuration}

This section describes the experimental setup, including the libraries, parameters of the methods and classification algorithms used in this study.

\subsubsection{Libraries}

Our feature extraction methods were implemented using PyTorch~\cite{Paszke2019PyTorch:Libraryb}, a popular open-source machine learning library. This platform is especially beneficial due to its application for training and manipulating deep learning models. To investigate the performance of some backbones on our dataset, we employed Pytorch Image Models (TIMM)~\cite{Wightman2019PyTorchtimm}, which includes a wide-range of pre-trained computer vision backbones, facilitating the implementation of different architectures. The classification process was realized through the use of the SVM implementation in
the Scikit-learn \cite{Pedregosa2011Scikit-learn:Python} library.

\subsubsection{Backbones}

We experimented with four different pre-trained backbones for our feature extraction methods. Specifically, we used ResNets~\cite{resnet} (18 and 50), both of which provide five distinct depths for feature extraction.
In addition to ResNets, we also considered two variants of the ConvNext architecture \cite{convnext}: ConvNext-Large-In22k and ConvNext-Xlarge-In22k. Both are pre-trained on the ImageNet-22k dataset, and have four different feature extraction depths. To further detail the complexity of each backbone, Table~\ref{tab:backboneparams} outlines their number of parameters and GFLOPs.

\begin{table}[!htpb]
\centering
\caption{Number of parameters for each backbone and respective GFLOPs used in this paper.}
\begin{tabular}{c|c|c}
\hline
\textbf{Backbone} & \textbf{N. of Params (millions)}  & \textbf{GFLOPs}\\
\hline
ResNet-18 & ~11.7 & 1.8 \\ 
\hline
ResNet-50 & ~25.5 & 4.1\\ 
\hline
ConvNext Large & ~230 & 34.4 \\
\hline
ConvNext Xlarge & ~392 & 60.9 \\
\hline
\end{tabular}
\label{tab:backboneparams}
\end{table}

\subsubsection{RADAM and Feature Extraction Configuration}

The feature extraction methods in this study are applied to images with a $224\times 224$ resolution to reduce the computational cost. In addition, the RADAM method incorporates the use of 4 RAEs, following the configuration suggested by the authors~\cite{scabini2023}, which provides a balance between performance and computational cost.

The performance of the different backbones was evaluated across three combinations of anatomical sections: Transversal only, Transversal + Tangential and Transversal + Tangential + Radial. Moreover, for each of these section combinations, we also compared the performance of both feature fusion strategies described in 
Section~\ref{sec:featureExtraction}, i.e., Serial Feature Concatenation (SFC) and Parallel Feature Merging (PFM).

\subsubsection{Classification Process}
The SVM was used as the primary classification algorithm for this study. We use a linear kernel and $C=1$, and no hyperparameter tuning was done. For each of the datasets detailed in Section~\ref{sec:datasets}, we adopted a 10-fold cross-validation strategy
and repeated the classification process 10 times to ensure a comprehensive evaluation of the SVM's performance across datasets.
Furthermore, we computed the average accuracy and its standard deviation over the 10 repetitions. These were the key metrics for assessing the performance of our proposed models. This provides a balanced view of the model's performance, taking into account both the model's correctness and consistency across different runs.

\section{Results}

In this section, we present and discuss the results obtained with each of the two feature extraction models.

\subsection{GAP Results}

In the first phase of our investigation with GAP, we explored our approach by analyzing the accuracies achieved for each individual anatomical section, for all datasets and all backbones. This examination aims to provide detailed insights into the strength and performance of GAP when applied to separate anatomical sections. It further aids in identifying the contribution of each section towards the overall accuracy in species identification. We present the results for each anatomical section (Transversal (Tr), tangential (T) and radial (R)) in Table~\ref{tab:results1_isolated}, across all four datasets: Original ($1000\times1000$), $1000\times500$, $500\times500$ and $500\times500$-OGRN and for each backbone in Table~\ref{tab:backboneparams}.

\begin{table}[!htpb]
\centering
\caption{Accuracies (expressed as percentages) and standard deviations of SVM using GAP features for each individual anatomical section, dataset and backbone.}
\resizebox{\textwidth}{!}{
\begin{tabular}{l|c*{4}{c}}
\toprule
 & & \multicolumn{4}{c}{Backbone} \\
\cmidrule(lr){3-6}
Dataset & Section & {Resnet18} & {Resnet50} & {Convnext Large} & {Convnext XLarge} \\
\midrule
\multirow{6}{*}{} Original & \multicolumn{1}{r}{Tr} &  67.00\tiny$\pm{0.90}$ & 61.70\tiny$\pm{0.50}$ & 79.90\tiny$\pm{1.10}$ & 82.20\tiny$\pm{1.10}$ \\  
& \multicolumn{1}{r}{T} & 74.50\tiny$\pm{0.60}$ & 67.30\tiny$\pm{0.80}$ & 87.30\tiny$\pm{0.80}$ & 88.20\tiny$\pm{0.60}$ \\
 & \multicolumn{1}{r}{R} & 57.30\tiny$\pm{0.90}$ & 52.40\tiny$\pm{0.70}$ & 75.40\tiny$\pm{1.20}$ & 77.60\tiny$\pm{0.80}$ \\
\midrule
\multirow{6}{*}{} $1000\times500$ & \multicolumn{1}{r}{Tr} &  85.70\tiny$\pm{0.50}$ & 82.70\tiny$\pm{0.40}$ & 96.10\tiny$\pm{0.30}$ & 96.30\tiny$\pm{0.40}$ \\ 
 & \multicolumn{1}{r}{T} & 82.60\tiny$\pm{0.40}$ & 82.20\tiny$\pm{0.20}$ & 95.00\tiny$\pm{0.60}$ & 95.40\tiny$\pm{0.40}$ \\
 & \multicolumn{1}{r}{R} & 75.30\tiny$\pm{0.50}$ & 73.30\tiny$\pm{0.40}$ & 91.00\tiny$\pm{0.40}$ & 91.50\tiny$\pm{0.40}$ \\ 
\midrule
\multirow{6}{*}{} $500\times500$ & \multicolumn{1}{r}{Tr} &  91.00\tiny$\pm{0.30}$ & 90.20\tiny$\pm{0.40}$ & 98.10\tiny$\pm{0.10}$ & 98.70\tiny$\pm{0.20}$ \\ 
 & \multicolumn{1}{r}{T} & 88.30\tiny$\pm{0.40}$ & 89.60\tiny$\pm{0.30}$ & 97.40\tiny$\pm{0.20}$ & 97.50\tiny$\pm{0.10}$ \\
 & \multicolumn{1}{r}{R} & 76.90\tiny$\pm{0.30}$ & 79.00\tiny$\pm{0.20}$ & 93.10\tiny$\pm{0.20}$ & 92.90\tiny$\pm{0.30}$ \\

\midrule
\multirow{6}{*}{} $500\times500$-OGRN & \multicolumn{1}{r}{Tr} &  78.40\tiny$\pm{0.50}$ & 74.90\tiny$\pm{0.80}$ & 91.50\tiny$\pm{0.40}$ & 92.60\tiny$\pm{0.50}$\\ 
 & \multicolumn{1}{r}{T} & 79.60\tiny$\pm{0.70}$ & 78.50\tiny$\pm{0.30}$ & 91.50\tiny$\pm{0.50}$ & 93.50\tiny$\pm{0.40}$ \\
 & \multicolumn{1}{r}{R} & 60.60\tiny$\pm{0.50}$ & 61.70\tiny$\pm{0.50}$ & 82.80\tiny$\pm{0.50}$ & 83.70\tiny$\pm{0.50}$ \\
\bottomrule
\end{tabular}
}
\label{tab:results1_isolated}
\end{table}

Our results indicate that the Tr and T anatomical sections provide more discriminative power for GAP in comparison to the radial (R) section. This could be attributed to the unique cellular arrangement and structures visible in the transversal and tangential sections, which may provide more distinctive features for the identification process. The radial section, while still providing valuable insights, seems to exhibit a less discriminative performance, possibly due to its inherent similarity among various species.

\begin{figure}[!htpb]
    \centering
    \includegraphics[width=0.8\textwidth]{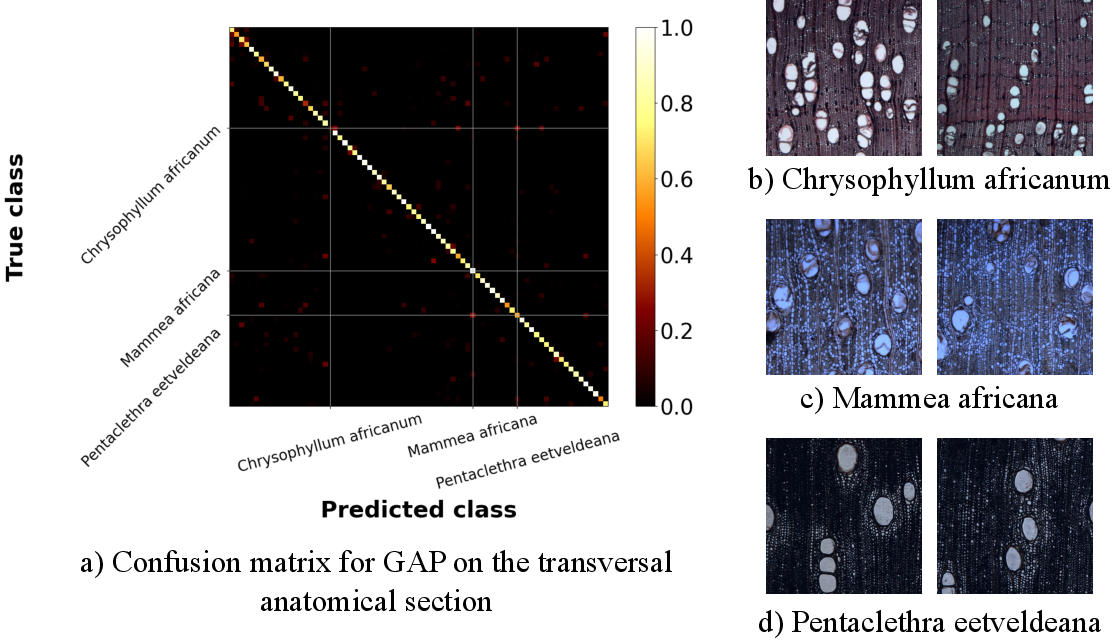}
    \caption{Confusion matrix for the GAP results on the transversal anatomical section in the original dataset. Class b) \textit{Chrysophyllum africanum }is often misclassified with d) \textit{Pentaclethra eetveldeana}, which is also confused with c) \textit{Mammea africana} class.}
    \label{fig:cm-tr-gap}
\end{figure}

There are also some interesting insights that can be derived from the classification results of the individual anatomical sections. 
Figure~\ref{fig:cm-tr-gap} displays
the confusion matrix of the results for the transversal anatomical section. Class b) \textit{Chrysophyllum africanum} tends to be misidentified as d) \textit{Pentaclethra eetveldeana}, which is also misclassified with c) \textit{Mammea africana}. This highlights the challenge of distinguishing between these species in the transversal plane, suggesting a need for more discriminative features for these species. 
Many features commonly identified as diagnostic are usually observed on a radial section, often requiring higher microscopic magnification for clear visibility. These characteristics are not easily distinguishable in the available images captured at lower magnifications. 

\begin{figure}[!htpb]
    \centering
    \includegraphics[width=0.8\textwidth]{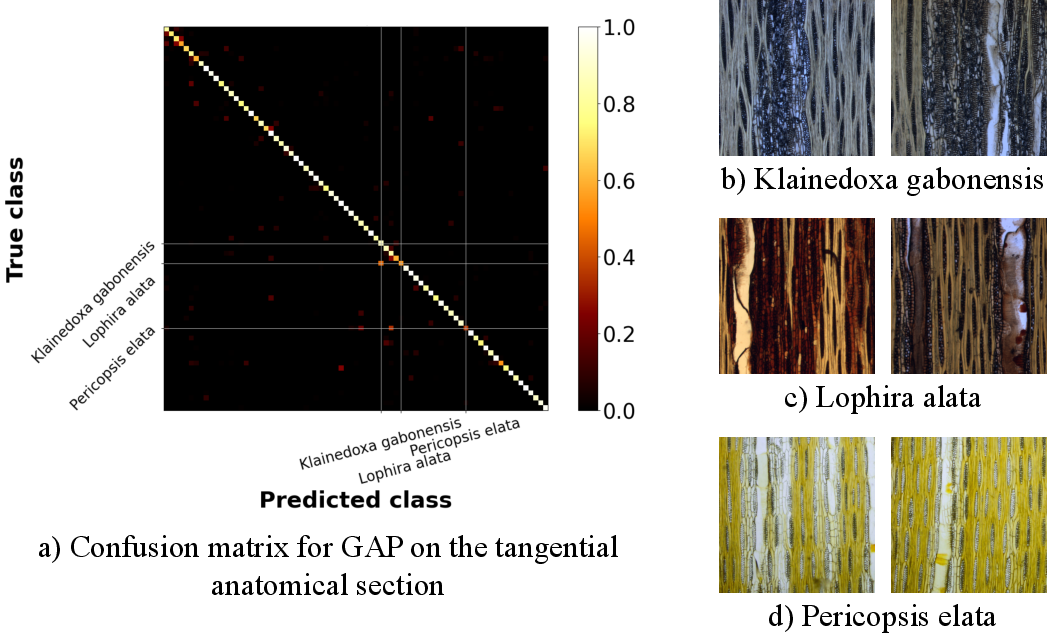}
    \caption{Confusion matrix for the GAP results on the tangential anatomical section in the original dataset. The model often misclassifies c) \textit{Lophira alata} with b) \textit{Klainedoxa gabonensis}, which is in general correctly classified. Also, class d) \textit{Pericopsis elata} is often confused with other several species.}
    \label{fig:cm-tg-gap}
\end{figure}

Proceeding to the results for the tangential anatomical section shown in Figure~\ref{fig:cm-tg-gap}, one can note that the model frequently confuses b) \textit{Klainedoxa gabonensis} and c) \textit{Lophira alata}. Another class, d) \textit{Pericopsis elata}, often finds itself misclassified across various species.

\begin{figure}[!htpb]
    \centering
    \includegraphics[width=0.8\textwidth]{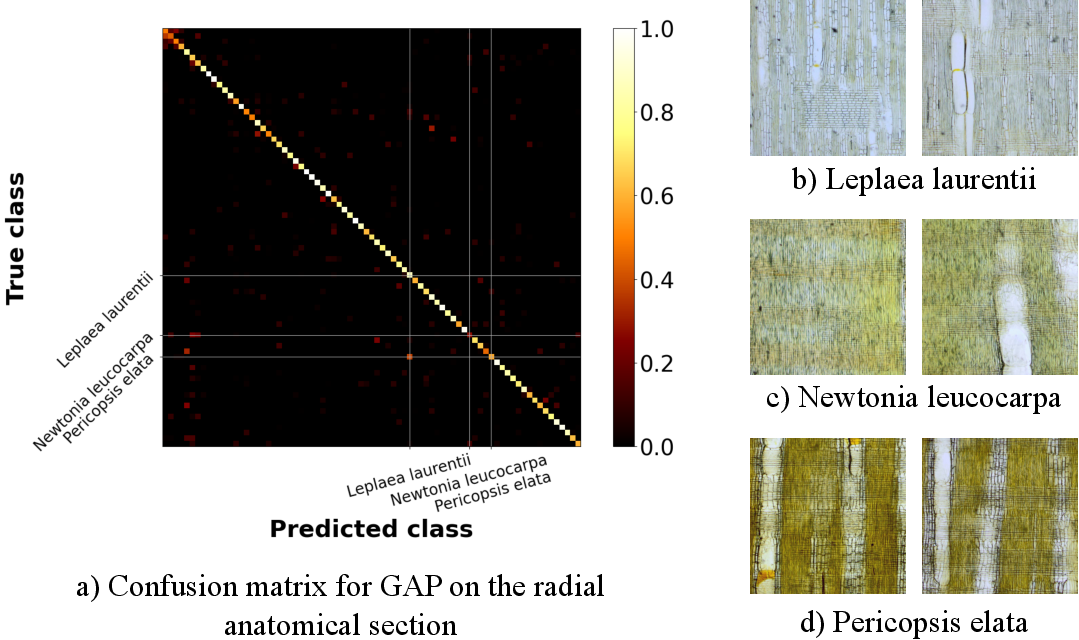}
    \caption{Confusion matrix for the GAP results on the radial anatomical section in the original dataset. Class b) \textit{Leplaea laurentii} is in general correctly classified, but d) \textit{Pericopsis elata} is frequently confused with \textit{Leplaea laurentii}. Also, d) \textit{Newtonia leucocarpa} has a very low classification accuracy.}
    \label{fig:cm-r-gap}
\end{figure}

Finally, for the radial anatomical section as represented by Figure~\ref{fig:cm-r-gap}, b) \textit{Leplaea laurentii} generally tends to be correctly classified, indicating effective differentiation by the model in this case. However, d) \textit{Pericopsis elata} is frequently misidentified as \textit{Leplaea laurentii}, indicating a gap in the model's discriminatory power between these two classes. Also, c) \textit{Newtonia leucocarpa} has a significantly low classification accuracy, reflecting its challenging identification in the radial plane.

Building upon these observations, we further analyze the use of combinations of anatomical sections, specifically transversal with tangential (Tr + T) and also all sections combined (Tr + T + R). Our goal is to investigate whether the complementary information from multiple sections can enhance the accuracy, and if so, to what extent. Incorporating more than one anatomical section allows the model to harness a broader spectrum of wood features, potentially improving the overall classification performance. The corresponding results are listed in Table~\ref{table:results1_combined}.

\begin{table}[!htbp]
\centering
\caption{Accuracies (expressed as percentages) and standard deviations of SVM using GAP features for different combinations of anatomical sections, and each fusion strategy, dataset and backbone.}
\label{table:results1_combined}
\resizebox{\textwidth}{!}{
\begin{tabular}{l|l|l|*{4}{c}}
\toprule
 & & & \multicolumn{4}{c}{Backbone} \\
\cmidrule(lr){4-7}
Dataset & Combination & Fusion Strategy & {Resnet18} & {Resnet50} & {Convnext Large} & {Convnext XLarge} \\
\midrule
\multirow{6}{*}{} Original &  \multicolumn{1}{r}{Tr+T} & \multicolumn{1}{r}{SFC} & 85.70\tiny$\pm{0.50}$ & 76.90\tiny$\pm{0.60}$ & 91.20\tiny$\pm{0.70}$ & 91.80\tiny$\pm{0.70}$ \\
& \multicolumn{1}{r}{} & \multicolumn{1}{r}{PFM} & 84.70\tiny$\pm{0.80}$ & 78.30\tiny$\pm{0.60}$ & 90.5\tiny$\pm{0.60}$ & 91.30\tiny$\pm{0.40}$ \\ \cline{2-7}\noalign{\vskip 1mm}
 & \multicolumn{1}{r}{Tr+T+R} & \multicolumn{1}{r}{SFC} & 87.40\tiny$\pm{0.90}$ & 80.80\tiny$\pm{0.50}$ & 92.50\tiny$\pm{0.50}$ & 93.10\tiny$\pm{0.70}$ \\
& \multicolumn{1}{r}{} & \multicolumn{1}{r}{PFM} & 86.40\tiny$\pm{0.90}$ & 83.0 \tiny$\pm{0.50}$ & 92.10\tiny$\pm{0.60}$ & 93.20\tiny$\pm{0.60}$ \\ 
\midrule
\multirow{6}{*}{} $1000\times500$ & \multicolumn{1}{r}{Tr+T} & \multicolumn{1}{r}{SFC} & 95.10\tiny$\pm{0.50}$ & 93.00\tiny$\pm{0.50}$ & 98.90\tiny$\pm{0.20}$ & 99.00\tiny$\pm{0.30}$ \\
& \multicolumn{1}{r}{} & \multicolumn{1}{r}{PFM} & 92.80\tiny$\pm{0.60}$ & 93.20\tiny$\pm{0.50}$ & 98.70\tiny$\pm{0.40}$ & 98.60\tiny$\pm{0.40}$ \\ \cline{2-7}\noalign{\vskip 1mm}
 & \multicolumn{1}{r}{Tr+T+R} & \multicolumn{1}{r}{SFC} & 96.80\tiny$\pm{0.40}$ & 94.50\tiny$\pm{0.40}$ & 99.20\tiny$\pm{0.30}$ & 99.30\tiny$\pm{0.20}$ \\
& \multicolumn{1}{r}{} & \multicolumn{1}{r}{PFM} & 95.00\tiny$\pm{0.40}$ & 94.90\tiny$\pm{0.40}$ & 98.80\tiny$\pm{0.20}$ & 98.90\tiny$\pm{0.30}$ \\ 
\midrule
\multirow{6}{*}{} $500\times500$ & \multicolumn{1}{r}{Tr+T} & \multicolumn{1}{r}{SFC} & 97.80\tiny$\pm{0.10}$ & 97.50\tiny$\pm{0.10}$ & 99.80\tiny$\pm{0.10}$ & 99.80\tiny$\pm{0.00}$ \\
& \multicolumn{1}{r}{} & \multicolumn{1}{r}{PFM} & 96.00\tiny$\pm{0.20}$ & 97.00\tiny$\pm{0.10}$ & 99.50\tiny$\pm{0.10}$ & 99.60\tiny$\pm{0.10}$ \\ \cline{2-7}\noalign{\vskip 1mm}
 & \multicolumn{1}{r}{Tr+T+R} & \multicolumn{1}{r}{SFC} & 97.20\tiny$\pm{0.30}$ & 98.30\tiny$\pm{0.20}$ & 99.80\tiny$\pm{0.00}$ & 99.80\tiny$\pm{0.10}$ \\
& \multicolumn{1}{r}{} & \multicolumn{1}{r}{PFM} & 97.00\tiny$\pm{0.20}$ & 97.70\tiny$\pm{0.10}$ & 99.60\tiny$\pm{0.10}$ & 99.70\tiny$\pm{0.10}$ \\ 
\midrule
\multirow{6}{*}{} $500\times500$-OGRN &  \multicolumn{1}{r}{Tr+T} & \multicolumn{1}{r}{SFC} & 92.40\tiny$\pm{0.40}$ & 89.5\tiny$\pm{0.40}$ & 97.20\tiny$\pm{0.30}$ & 97.90\tiny$\pm{0.20}$ \\
& \multicolumn{1}{r}{} & \multicolumn{1}{r}{PFM} & 89.60\tiny$\pm{0.50}$ & 89.80\tiny$\pm{0.50}$ & 96.30\tiny$\pm{0.30}$ & 97.50\tiny$\pm{0.20}$ \\ \cline{2-7}\noalign{\vskip 1mm}
 & \multicolumn{1}{r}{Tr+T+R} & \multicolumn{1}{r}{SFC} & 93.40\tiny$\pm{0.40}$ & 92.50\tiny$\pm{0.50}$ & 97.70\tiny$\pm{0.20}$ & 98.20\tiny$\pm{0.20}$ \\
& \multicolumn{1}{r}{} & \multicolumn{1}{r}{PFM} & 91.10\tiny$\pm{0.40}$ & 92.50\tiny$\pm{0.50}$ & 97.00\tiny$\pm{0.30}$ & 97.90\tiny$\pm{0.20}$ \\ 
\bottomrule
\end{tabular}
}
\end{table}

We start the discussion of the results with the performance of the model on the original dataset, which has fewer images per class compared to the other datasets. As the results suggest, this dataset yielded the lowest accuracy among all datasets, with the highest accuracy of $93.2\pm{0.6}$ obtained when all the anatomical sections are considered using the PFM fusion strategy and the ConvNext XLarge backbone.

In contrast, the accuracy of the model dramatically improved when we increased the number of images per class in the datasets. For instance, on the $1000\times500$ dataset, it reached an accuracy of $99.3\pm{0.2}$ (all anatomical sections, SFC fusion strategy and ConvNext XLarge backbone), and on the $500\times500$ dataset, it reached an even higher performance of $99.8\pm{0.0}$ (all anatomical sections, SFC fusion strategy
and ConvNetx Large). Notably, on the $500\times500$-OGRN dataset, which contains the same number of images but incorporates noise,
the model maintained a high accuracy of $98.2\pm{0.2}$ when using the ConvNext XLarge backbone. This result underlines the observation that increasing the number of samples per class in a dataset significantly enhances the performance of the model.

Another factor that has a positive impact on the model performance is the inclusion of more anatomical sections in the analysis. This is clear from the results obtained on the original dataset using the ConvNext Large backbone. When we considered the transversal section (see Table~\ref{tab:results1_isolated}) and the ResNet18 backbone, the accuracy was $67.0\pm{0.90}$. However, the accuracy increased to $85.7\pm{0.5}$ upon the inclusion of the tangential section and further increased to $87.4\pm{0.8}$ when also the radial section was incorporated.

In terms of feature fusion strategies, one can see that SFC generally outperforms PFM. However, it should be noted that the SFC strategy results in a larger feature vector due to concatenation, which may, sometimes, be a disadvantage as it can reduce the efficiency of the classification step. On the other hand, PFM allows for a better control of the number of features regardless of the number of terms involved in the summations.

Finally, the comparison of different backbones reveals some interesting insights. Although ResNet18 generally outperforms ResNet50, especially when using SFC, the ConvNext backbones display superior performance across all datasets, regardless of the combination of sections and the fusion strategy considered. Given that they have a considerably larger size, better pre-training and a more advanced architecture design compared to the ResNets, this was an expected result.

\subsection{RADAM Results}

In this section, we detail the results obtained with the RADAM feature extraction method. Similar to GAP, we evaluate the performance across different datasets starting with the analysis of individual anatomical sections. The results are listed in Table~\ref{tab:results2_isolated}.

\begin{table}[!htpb]
\centering
\caption{Accuracies (expressed as percentages) and standard deviations of SVM using RADAM features for each individual anatomical setion, dataset and backbone.}
\resizebox{\textwidth}{!}{
\begin{tabular}{l|c*{4}{c}}
\toprule
 & & \multicolumn{4}{c}{Backbone} \\
\cmidrule(lr){3-6}
Dataset & Section & {Resnet18} & {Resnet50} & {Convnext Large} & {Convnext XLarge} \\
\midrule
\multirow{6}{*}{} Original & \multicolumn{1}{r}{Tr} &  77.00\tiny$\pm{0.50}$ & 77.20\tiny$\pm{0.70}$ & 86.10\tiny$\pm{0.70}$ & 87.30\tiny$\pm{0.70}$ \\  
& \multicolumn{1}{r}{T} & 82.00\tiny$\pm{0.40}$ & 83.70\tiny$\pm{0.70}$ & 90.30\tiny$\pm{0.90}$ & 90.50\tiny$\pm{0.60}$ \\
 & \multicolumn{1}{r}{R} & 69.70\tiny$\pm{0.80}$ & 68.90\tiny$\pm{0.60}$ & 81.70\tiny$\pm{0.90}$ & 83.60\tiny$\pm{0.90}$ \\
\midrule
\multirow{6}{*}{} $1000\times500$ & \multicolumn{1}{r}{Tr} &  94.80\tiny$\pm{0.40}$ & 94.30\tiny$\pm{0.50}$ & 97.90\tiny$\pm{0.30}$ & 98.30\tiny$\pm{0.30}$ \\ 
 & \multicolumn{1}{r}{T} & 91.00\tiny$\pm{0.50}$ & 91.90\tiny$\pm{0.30}$ & 96.90\tiny$\pm{0.30}$ & 96.70\tiny$\pm{0.50}$ \\
 & \multicolumn{1}{r}{R} & 88.80\tiny$\pm{0.40}$ & 88.70\tiny$\pm{0.70}$ & 95.00\tiny$\pm{0.30}$ & 95.30\tiny$\pm{0.30}$ \\ 
\midrule
\multirow{6}{*}{} $500\times500$ & \multicolumn{1}{r}{Tr} &  97.30\tiny$\pm{0.20}$ & 97.50\tiny$\pm{0.20}$ & 99.10\tiny$\pm{0.10}$ & 99.40\tiny$\pm{0.10}$ \\ 
 & \multicolumn{1}{r}{T} & 95.50\tiny$\pm{0.10}$ & 96.70\tiny$\pm{0.20}$ & 99.00\tiny$\pm{0.10}$ & 98.90\tiny$\pm{0.10}$ \\
 & \multicolumn{1}{r}{R} & 90.80\tiny$\pm{0.30}$ & 91.90\tiny$\pm{0.20}$ & 96.80\tiny$\pm{0.20}$ & 96.20\tiny$\pm{0.10}$ \\

\midrule
\multirow{6}{*}{} $500\times500$-OGRN & \multicolumn{1}{r}{Tr} &  87.50\tiny$\pm{0.20}$ & 87.60\tiny$\pm{0.40}$ & 94.70\tiny$\pm{0.30}$ & 94.80\tiny$\pm{0.40}$\\ 
 & \multicolumn{1}{r}{T} & 88.20\tiny$\pm{0.50}$ & 90.00\tiny$\pm{0.30}$ & 95.50\tiny$\pm{0.30}$ & 96.40\tiny$\pm{0.30}$ \\
 & \multicolumn{1}{r}{R} & 75.30\tiny$\pm{0.50}$ & 78.10\tiny$\pm{0.50}$ & 88.70\tiny$\pm{0.30}$ & 88.70\tiny$\pm{0.60}$ \\
\bottomrule
\end{tabular}
}
\label{tab:results2_isolated}
\end{table}

The results obtained with RADAM mirror the previous observations for GAP. Both the Tr and T sections appear to have stronger discriminative capabilities compared to the R section. The superior performance of these sections may be attributed to their distinctive cellular structures and arrangements that RADAM effectively captures, thereby providing more characteristic features for the wood species identification task.

\begin{figure}[!htpb]
    \centering
    \includegraphics[width=0.8\textwidth]{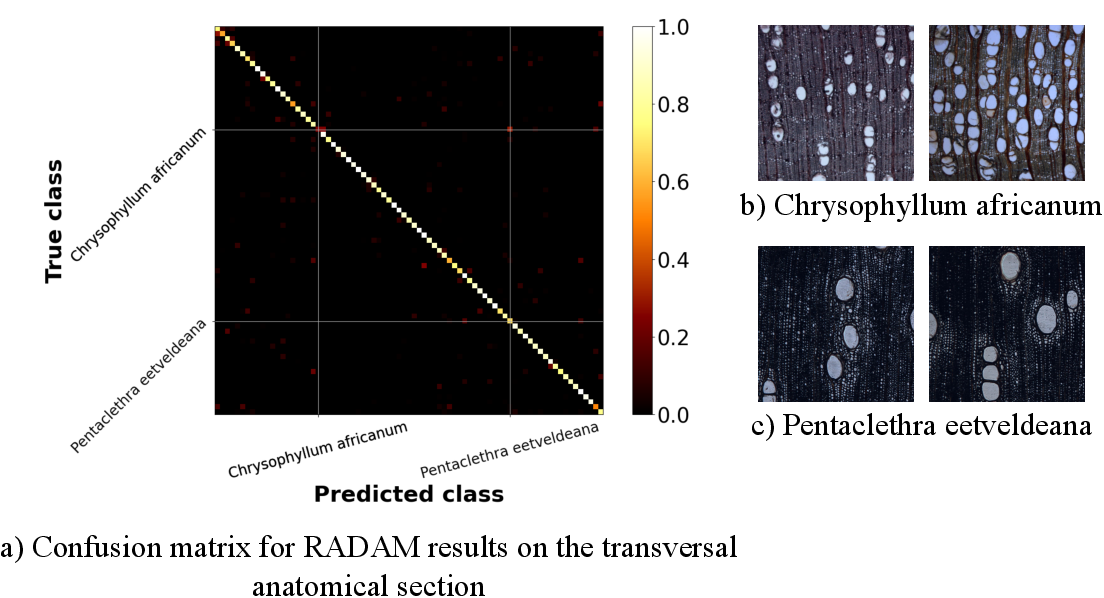}
    \caption{Confusion matrix for the RADAM results on the transversal anatomical section in the original dataset. Class b) \textit{Chrysophyllum africanum} has a low classification accuracy, and is frequently confused with class c) \textit{Pentaclethra eetveldeana}.}
    \label{fig:cm-tr-radam}
\end{figure}

We also made a deeper investigation for each anatomical section by inspecting its confusion matrix, as we did for GAP. Figure~\ref{fig:cm-tr-radam} presents the confusion matrix for the RADAM results on the transversal anatomical section in the original dataset. Class b) \textit{Chrysophyllum africanum} has a low classification accuracy and is often confused with class c) \textit{Pentaclethra eetveldeana}. This confusion reflects the model's struggle to distinctly identify these two classes for this particular section.

For the tangential anatomical section, as shown in Figure~\ref{fig:cm-tr-radam}, class b) \textit{Leplaea laurentii} is generally accurately identified by the model. However, a noticeable confusion emerges with class c) \textit{Pericopsis elata}, which is frequently misclassified as \textit{Leplaea laurentii}. 

\begin{figure}[!htpb]
    \centering
    \includegraphics[width=0.8\textwidth]{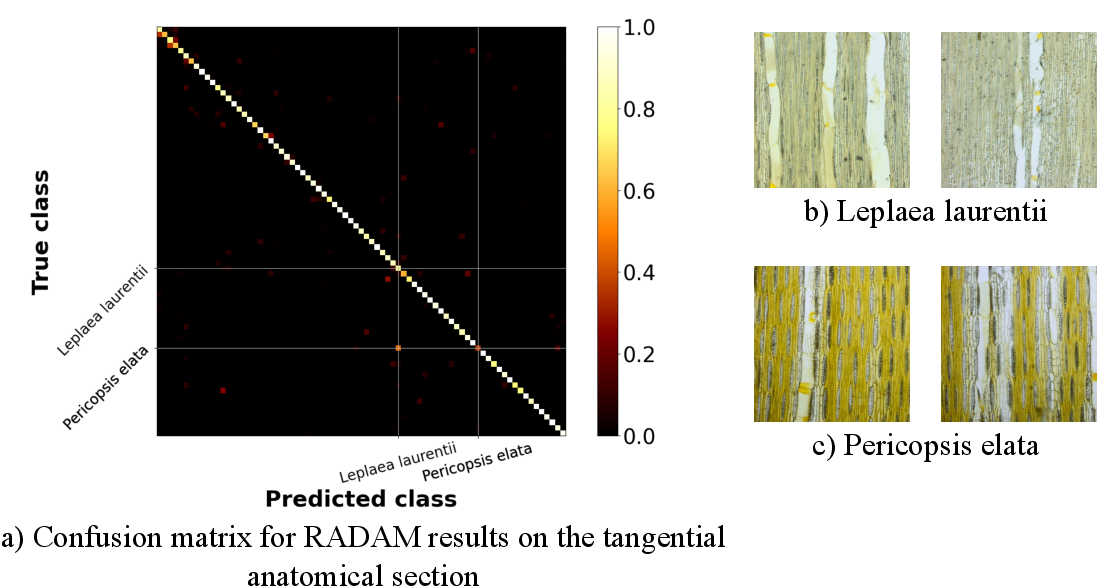}
    \caption{Confusion matrix for the RADAM results on the tangential anatomical section in the original dataset. Class b) \textit{Leplaea laurentii} is in general correctly classified, but it is also confused with class c) \textit{Pericopsis elata}.}
    \label{fig:cm-tg-radam}
\end{figure}

\begin{figure}[!htpb]
    \centering
    \includegraphics[width=0.8\textwidth]{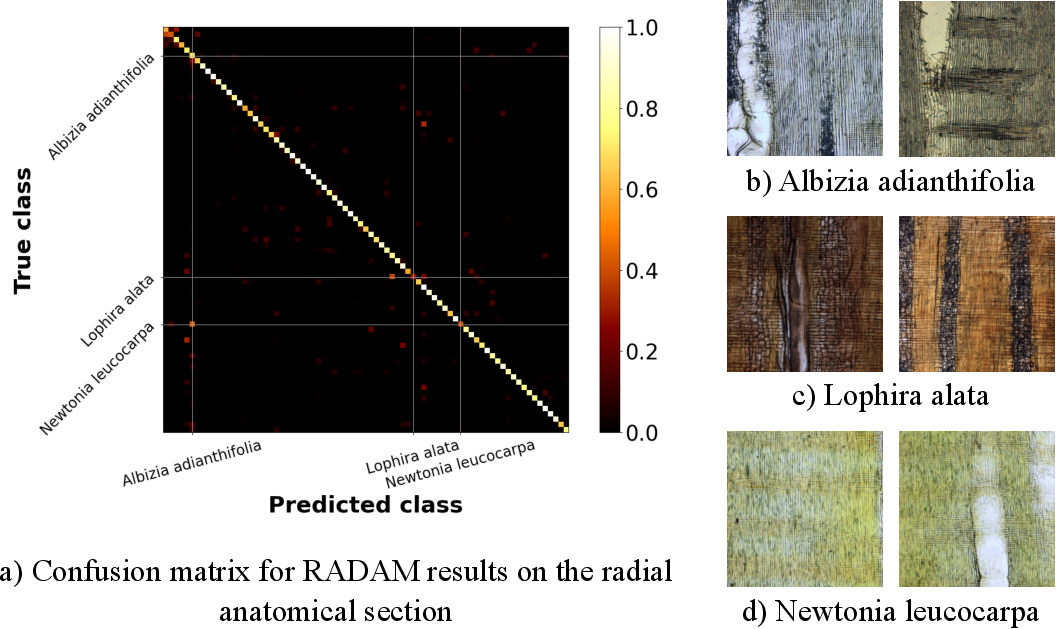}
    \caption{Confusion matrix for the RADAM results on the radial anatomical section in the original dataset. Class a) \textit{Albizia adianthifolia} is well distinguished, but it is also confused with class d) \textit{Newtonia leucocarpa}. Also, class c) \textit{Lophira alata} is misclassified with other many species.}
    \label{fig:cm-r-radam}
\end{figure}

Finally, for the radial anatomical section, as shown in Figure~\ref{fig:cm-r-radam}, class b) \textit{Albizia adianthifolia} is well distinguished by the model. Yet, it is often misclassified as
class d) \textit{Newtonia leucocarpa}, which shows a significant confusion with \textit{Albizia adianthifolia}. Furthermore, class c) \textit{Lophira alata} is often misclassified as one of various other species. This implies that the model encounters difficulties in accurately classifying these species.

These observations led us to further investigate the combination of anatomical sections, specifically Tr + T and Tr + T + R, and also testing different feature fusion strategies, with the intention of complementing information from different sections to improve the classification accuracy. The results are shown in Table~\ref{table:results2_combined}.

\begin{table}[!htbp]
\centering
\caption{Accuracies (expressed as percentages) and standard deviations of SVM using RADAM features for different combinations of anatomical sections, and each fusion strategy, dataset and backbone.}
\label{table:results2_combined}
\resizebox{\textwidth}{!}{
\begin{tabular}{l|l|l|*{4}{c}}
\toprule
 & & & \multicolumn{4}{c}{Backbone} \\
\cmidrule(lr){4-7}
Dataset & Combination & Fusion Strategy & {Resnet18} & {Resnet50} & {Convnext Large} & {Convnext XLarge} \\
\midrule
\multirow{6}{*}{} Original &\multicolumn{1}{r}{Tr+T} & \multicolumn{1}{r}{SFC} & 90.20\tiny$\pm{0.70}$ & 89.00\tiny$\pm{0.60}$ & 93.50\tiny$\pm{0.50}$ & 94.00\tiny$\pm{0.60}$ \\
& \multicolumn{1}{r}{} & \multicolumn{1}{r}{PFM} & 88.60\tiny$\pm{0.70}$ & 88.20\tiny$\pm{0.60}$ & 92.60\tiny$\pm{0.50}$ & 93.60\tiny$\pm{0.50}$ \\ \cline{2-7}\noalign{\vskip 1mm}
 & \multicolumn{1}{r}{Tr+T+R} & \multicolumn{1}{r}{SFC} & 91.20\tiny$\pm{0.60}$ & 90.60\tiny$\pm{0.40}$ & 94.00\tiny$\pm{0.70}$ & 94.30\tiny$\pm{0.70}$ \\
& \multicolumn{1}{r}{} & \multicolumn{1}{r}{PFM} & 90.90\tiny$\pm{0.80}$ & 89.60\tiny$\pm{0.80}$ & 94.70\tiny$\pm{0.60}$ & 94.80\tiny$\pm{0.60}$ \\ 
\midrule
\multirow{6}{*}{} $1000\times500$ & \multicolumn{1}{r}{Tr+T} & \multicolumn{1}{r}{SFC} & 98.10\tiny$\pm{0.40}$ & 98.00\tiny$\pm{0.30}$ & 99.40\tiny$\pm{0.30}$ & 99.30\tiny$\pm{0.30}$ \\
& \multicolumn{1}{r}{} & \multicolumn{1}{r}{PFM} & 97.70\tiny$\pm{0.30}$ & 97.90\tiny$\pm{0.40}$ & 99.20\tiny$\pm{0.30}$ & 99.20\tiny$\pm{0.30}$ \\ \cline{2-7}\noalign{\vskip 1mm}
 & \multicolumn{1}{r}{Tr+T+R} & \multicolumn{1}{r}{SFC} & 98.50\tiny$\pm{0.40}$ & 98.60\tiny$\pm{0.30}$ & 99.60\tiny$\pm{0.20}$ & 99.50\tiny$\pm{0.20}$ \\
& \multicolumn{1}{r}{} & \multicolumn{1}{r}{PFM} & 97.70\tiny$\pm{0.40}$ & 98.10\tiny$\pm{0.40}$ & 99.40\tiny$\pm{0.20}$ & 99.40\tiny$\pm{0.20}$ \\ 
\midrule
\multirow{6}{*}{} $500\times500$ &\multicolumn{1}{r}{Tr+T} & \multicolumn{1}{r}{SFC} & 99.40\tiny$\pm{0.10}$ & 99.50\tiny$\pm{0.10}$ & 99.90\tiny$\pm{0.00}$ & 99.90\tiny$\pm{0.00}$ \\
& \multicolumn{1}{r}{} & \multicolumn{1}{r}{PFM} & 98.90\tiny$\pm{0.10}$ & 99.30\tiny$\pm{0.10}$ & 99.90\tiny$\pm{0.10}$ & 99.80\tiny$\pm{0.00}$ \\ \cline{2-7}\noalign{\vskip 1mm}
 & \multicolumn{1}{r}{Tr+T+R} & \multicolumn{1}{r}{SFC} & 99.50\tiny$\pm{0.10}$ & 99.60\tiny$\pm{0.10}$ & 99.90\tiny$\pm{0.00}$ & 99.90\tiny$\pm{0.10}$ \\
& \multicolumn{1}{r}{} & \multicolumn{1}{r}{PFM} & 99.10\tiny$\pm{0.10}$ & 99.30\tiny$\pm{0.10}$ & 99.80\tiny$\pm{0.00}$ & 99.80\tiny$\pm{0.00}$ \\ 
\midrule
\multirow{6}{*}{} $500\times500$-OGRN & \multicolumn{1}{r}{Tr+T} & \multicolumn{1}{r}{SFC} & 96.10\tiny$\pm{0.30}$ & 95.70\tiny$\pm{0.30}$ & 98.70\tiny$\pm{0.20}$ & 98.70\tiny$\pm{0.20}$ \\
& \multicolumn{1}{r}{} & \multicolumn{1}{r}{PFM} & 94.70\tiny$\pm{0.40}$ & 95.20\tiny$\pm{0.40}$ & 98.50\tiny$\pm{0.30}$ & 98.30\tiny$\pm{0.20}$ \\ \cline{2-7}\noalign{\vskip 1mm}
 & \multicolumn{1}{r}{Tr+T+R} & \multicolumn{1}{r}{SFC} & 96.30\tiny$\pm{0.30}$ & 96.20\tiny$\pm{0.30}$ & 98.70\tiny$\pm{0.20}$ & 98.70\tiny$\pm{0.20}$ \\
& \multicolumn{1}{r}{} & \multicolumn{1}{r}{PFM} & 95.40\tiny$\pm{0.40}$ & 95.60\tiny$\pm{0.40}$ & 98.50\tiny$\pm{0.30}$ & 98.60\tiny$\pm{0.20}$ \\ 
\bottomrule
\end{tabular}
}
\end{table}

We start the discussion of the results with the performance of the model on the original dataset, 
reaching its highest accuracy of $94.8\pm{0.5}$ when all the anatomical sections are considered using the PFM feature fusion strategy and ConvNext XLarge.

However, a remarkable performance improvement was observed as the number of images per class in the datasets increased. Specifically, the model accuracy increased to $99.50\pm{0.2}$ on the $1000\times500$ dataset (all anatomical sections, SFC strategy and 
ConvNext XLarge), while on the $500\times500$ dataset it reached a nearly perfect score of $99.9\pm{0.0}$ (multiple configurations) and finally on the $500\times500$-OGRN dataset the model maintained a good performance of $98.7\pm{0.2}$ (multiple configurations). These results, similar to the previous section, confirm the model's enhanced performance with an increase in the number of samples per class in the dataset.

Furthermore, the inclusion of more anatomical sections in the analysis also increased the accuracy of the RADAM model. For instance, considering only the transversal section in the original dataset using the ResNet18 backbone yielded an accuracy of $77.0\pm{0.5}$
(see Table~\ref{tab:results2_isolated}, which increased to $90.2\pm{0.7}$
when including the tangential section, and further improved to $91.2\pm{0.6}$ when incorporating the radial section, using the SFC strategy.

As for the feature fusion strategies, in general, SFC continues to be better than PFM, 
but we can note that there are some cases, for example, the ConvNext Large backbone on the original dataset with all anatomical sections, where PFM outperforms SFC. 

Lastly, the results show a similar comparison of the backbones as in the previous section, where in many cases ResNet18 outperforms ResNet50, but in general the ConvNexts delivered the best performance across all datasets regardless of the combination of 
sections and the fusion strategy considered.

\subsection{Overall comparison}

In this subsection, we present the main results for the four distinct datasets, taking into account different combinations of anatomical sections for the methods introduced in this study, as well as the approach proposed by Rosa da Silva et al~\cite{daSilva2022ImprovedPlanes}, which applies LPQ on each section, uses a random forest classifier on each section separately and further concatenates the probability matrices obtained in the classification process as input to a logistic regression model. A comprehensive comparison of these results is presented in Table~\ref{tab:comparison}.

\begin{table}[!htpb]
\centering
\caption{Accuracy (expressed as percentages) and standard deviations of the three methods for different datasets and different anatomical section combinations. For GAP and RADAM, we used the Convnext XLarge backbone and the SFC strategy when using more than one anatomical section.}
\label{tab:comparison}
\resizebox{\textwidth}{!}{
\begin{tabular}{c|ccccc}
\hline
\multirow{2}{*}{Dataset} & \multirow{2}{*}{Anatomical Section} & \multicolumn{3}{c}{Method}  \\ \cline{3-5} 
                          &                                     & GAP & RADAM & Rosa da Silva et al.~\cite{daSilva2022ImprovedPlanes} \\ \hline
Original                & Tr                                  &      82.2\tiny$\pm{1.10}$                       &   87.3\tiny$\pm{0.70}$    &  56.0\tiny$\pm{2.00}$                \\ 
                          & T                                   &           88.2\tiny$\pm{0.60}$                  &  90.5\tiny$\pm{0.60}$    & 42.0\tiny$\pm{2.00}$                  \\ 
                          & R                                   &          77.6\tiny$\pm{0.80}$                   &   83.6\tiny$\pm{0.90}$    & 42.0\tiny$\pm{2.00}$                  \\ 
                          & Tr + T                              &           91.8\tiny$\pm{0.70}$                  &   94.0\tiny$\pm{0.60}$    & 62.0\tiny$\pm{4.00}$                  \\ 
                          & Tr + T + R                          &         93.1\tiny$\pm{0.70}$                    &    94.3\tiny$\pm{0.70}$   & 66.0\tiny$\pm{2.00}$                  \\ \hline
$1000\times500$                & Tr                                  &          96.3\tiny$\pm{0.40}$                   &  98.3\tiny$\pm{0.30}$     &     71.0\tiny$\pm{2.00}$              \\ 
                          & T                                   &             95.4\tiny$\pm{0.40}$                & 96.7\tiny$\pm{0.50}$      & 71.0\tiny$\pm{1.00}$                  \\ 
                          & R                                   &         91.5\tiny$\pm{0.40}$                   &   95.3\tiny$\pm{0.30}$    & 52.0\tiny$\pm{1.00}$                  \\  
                          & Tr + T                              &     99.0\tiny$\pm{0.30}$                         &   99.3\tiny$\pm{0.30}$    & 85.0\tiny$\pm{2.00}$                  \\  
                          & Tr + T + R                          &    99.3\tiny$\pm{0.20}$                         &  99.5\tiny$\pm{0.20}$     & 91.0\tiny$\pm{2.00}$                  \\ \hline
$500\times500$                 & Tr                                  &    98.7\tiny$\pm{0.20}$                         &   99.4\tiny$\pm{0.10}$    &      75.0\tiny$\pm{2.00}$             \\  
                          & T                                   &    97.5\tiny$\pm{0.10}$                         &   98.9\tiny$\pm{0.10}$    & 69.0\tiny$\pm{1.00}$                  \\ 
                          & R                                   &    92.9\tiny$\pm{0.30}$                         &   96.2\tiny$\pm{0.10}$    & 54.0\tiny$\pm{1.00}$                  \\  
                          & Tr + T                              &     99.8\tiny$\pm{0.00}$                        &   99.9\tiny$\pm{0.00}$    & 86.0\tiny$\pm{2.00}$                  \\  
                          & Tr + T + R                          &     99.8\tiny$\pm{0.10}$                        &   99.9\tiny$\pm{0.10}$    & 95.0\tiny$\pm{1.00}$                  \\ \hline
$500\times500$-OGRN                & Tr                                  &     92.6\tiny$\pm{0.60}$                        &  94.8\tiny$\pm{0.40}$     &   38.0\tiny$\pm{2.00}$                \\ 
                          & T                                   &       93.5\tiny$\pm{0.40}$                      &  96.4\tiny$\pm{0.30}$     & 34.0\tiny$\pm{1.00}$                  \\  
                          & R                                   &      83.7\tiny$\pm{0.50}$                       &  88.7\tiny$\pm{0.60}$     & 27.0\tiny$\pm{1.00}$                  \\ 
                          & Tr + T                              &     97.9\tiny$\pm{0.20}$                        &   98.7\tiny$\pm{0.20}$    & 48.0\tiny$\pm{2.00}$                  \\  
                          & Tr + T + R                          &    98.2\tiny$\pm{0.20}$                         &   98.7\tiny$\pm{0.20}$    &62.0\tiny$\pm{3.00}$                   \\ \hline
\end{tabular}
}
\end{table}

An observation that can be made from our comparison is that both methods proposed in this study consistently surpass the performance of the approach by Rosa da Silva et al. \cite{daSilva2022ImprovedPlanes}, regardless of the dataset employed or the combination of anatomical sections considered. Furthermore, among our proposed methods, RADAM exhibits superior performance compared to GAP, demonstrating its robustness across different datasets, anatomical sections and feature fusion strategies.

\subsection{Leave-k-trees-out}

In our study, we also analyzed the performance of our method using the leave-k-trees-out cross-validation strategy, as used by Rosa da Silva et al.~\cite{daSilva2022ImprovedPlanes}. This cross-validation approach involves segregating all samples from a single tree for each species into the test set, thereby ensuring complete independence between the training and test sets in terms of specimens. Consequently, 165 samples were allocated for testing and 640 for training.

Table~\ref{tab:lktresults} presents a comparison among GAP, RADAM and the method proposed by Rosa da Silva et al.~\cite{daSilva2022ImprovedPlanes}. All methods were evaluated under the leave-k-trees-out cross-validation scheme. For our methods, we 
used all three anatomical sections and the SFC strategy.

\begin{table}
\caption{Accuracy comparison of GAP, RADAM and the method of Rosa da Silva et al~\cite{daSilva2022ImprovedPlanes}, analyzed with the leave-k-trees-out cross validation strategy. For our methods, we used all three anatomical sections and the SFC strategy.}
\begin{tabular}{l|cccc}
\toprule
{} & Original & $1000\times500$ & $500\times500$ & $500\times500$-OGRN \\
\midrule
Rosa da Silva et al.~\cite{daSilva2022ImprovedPlanes}        &  30.00       &     28.00           &         27.00       &          22.00          \\
GAP        &     92.17     &        91.87         &            95.03    &          93.96           \\
RADAM &     93.98     &     94.28            &        95.03        &       94.12              \\
\bottomrule
\end{tabular}
\label{tab:lktresults}
\end{table}

The results show a notable performance disparity. While the method of Rosa da Silva et al. exhibited accuracies ranging from 22.00\% to 30.00\% across different datasets, both GAP and RADAM demonstrated significantly higher accuracies, with RADAM slightly outperforming GAP. These results suggest that our method, although showing a slight inferior performance compared to the traditional cross-validation configuration, still surpasses the performance of Rosa da Silva et al.

In summary, the leave-k-trees-out cross-validation approach has demonstrated the robustness of GAP and RADAM in the context of significant sample variability when characterizing different trees/specimens. 

\section{Conclusions}
In this study, we investigated the efficacy of two feature extraction techniques (GAP and RADAM) used in conjunction with an SVM classifier for the purpose of wood species identification through texture analysis on images from three different anatomical sections. Our results indicate that SVM, when employing either GAP or RADAM, surpasses the results of Rosa da Silva et al.~\cite{daSilva2022ImprovedPlanes}, demonstrating their effectiveness in the context of wood species identification through texture analysis techniques. Our research also revealed that combining multiple anatomical sections can significantly enhance the performance.  

Among the two methods proposed in this study, RADAM consistently demonstrated superior performance across different datasets, anatomical section combinations and feature fusion strategies. This highlights the robustness of RADAM in handling diverse conditions and reinforces its potential as an effective tool, not only for the recognition of macroscopic textures, but also for wood species identification using microscopic images.

Moreover, our findings can potentially contribute to the automation of wood species identification, reducing the dependence on expert knowledge and enabling more efficient and scalable approaches. This is particularly relevant in the context of timber supply chain monitoring and regulation, where the ability to quickly and accurately identify wood species can facilitate interventions and promote responsible practices.

However, despite the promising results, future research may improve the characterization of different trees/specimens by further exploring the integration of other factors such as image resolution or the use of other neural network architectures such as Vision Transformers (ViTs).

In summary, the work proposed here represents a significant step forward in the field of wood species identification. Through exploiting the computational efficiency of pre-trained neural network models and advanced feature extraction techniques such as RADAM, we provided a robust and efficient approach to wood species identification, opening doors to more sustainable and responsible forestry practices.

\section*{Acknowledgements}

K. M. Zielinski acknowledges support from the São Paulo Research Foundation (FAPESP) (Grant \#2022/03668-1) and Higher Education Personnel Improvement Coordination (CAPES) (Grant  \#88887.631085/2021-00). L. Scabini acknowledges funding from FAPESP (Grants \#2021/09163-6 and \#2023/10442-2).  L. C. Ribas acknowledges support from FAPESP (grants \#2023/04583-2). O. M. Bruno acknowledges support from CNPq (Grant \#307897/2018-4) and FAPESP (grants \#2018/22214-6 and \#2021/08325-2). The authors are also grateful to the NVIDIA GPU Grant Program. B. De Baets  received funding from the Flemish Government under the "Onderzoeksprogramma Artificiële Intelligentie (AI) Vlaanderen" programme.

\bibliographystyle{elsarticle-num}
\bibliography{main}

\end{document}